\documentclass[11pt,a4paper]{article}
\usepackage[hyperref]{naaclhlt2019}
\usepackage{times}
\usepackage{latexsym}
\usepackage{helvet}  
\usepackage{courier}  
\usepackage{url}  
\usepackage{graphicx}  
\usepackage{amsfonts}
\usepackage{amsmath}
\usepackage{verbatim}
\usepackage{bm}
\usepackage{makecell}
\usepackage[font=small, labelfont=bf]{caption}
\usepackage[]{subcaption}
\usepackage{color, colortbl}
\usepackage[colorinlistoftodos]{todonotes}
\newcommand{\ourmethod}{\textsc{Mohone}}

\usepackage{url}

\aclfinalcopy 

\title{Instructions for NAACL-HLT 2019 Proceedings}

\author{First Author \\
  Affiliation / Address line 1 \\
  Affiliation / Address line 2 \\
  Affiliation / Address line 3 \\
  {\tt email@domain} \\\And
  Second Author \\
  Affiliation / Address line 1 \\
  Affiliation / Address line 2 \\
  Affiliation / Address line 3 \\
  {\tt email@domain} \\}

\date{}

\begin{document}
\title{\ourmethod: Modeling Higher Order Network Effects in Knowledge Graphs via Network Infused Embeddings}
\author{Hao Yu \\
    Department of Electronic Engineering \\ 
  	Tsinghua University, China \\
  {\tt haoy15@mails.tsinghua.edu.cn} \\ \And
  Vivek Kulkarni \\
   Department of Computer Science \\ 
   University of California, Santa Barbara \\
   {\tt vvkulkarni@cs.ucsb.edu} \\ \AND
  William Yang Wang \\
   Department of Computer Science \\ 
   University of California, Santa Barbara \\
  {\tt william@cs.ucsb.edu} \\
}
\maketitle
\begin{abstract}
Many knowledge graph embedding methods operate on triples and are therefore implicitly limited by a very local view of the entire knowledge graph. We present a new framework \textsc{\ourmethod} to effectively model higher order network effects in knowledge-graphs, thus enabling one to capture varying degrees of network connectivity (from the local to the global). Our framework is generic, explicitly models the network scale, and captures two different aspects of similarity in networks: (a) shared local neighborhood  and (b) structural role-based similarity. First, we introduce methods that learn network representations of entities in the knowledge graph capturing these varied aspects of similarity. We then propose a fast, efficient method to incorporate the information captured by these network representations into existing knowledge graph embeddings. We show that our method consistently and significantly improves the performance on link prediction of several different knowledge-graph embedding methods including \textsc{TransE, TransD, DistMult}, and \textsc{ComplEx} (by at least 4 points or  $17\%$ in some cases).

\end{abstract}


\section{Introduction}
Knowledge graphs \cite{dong2014knowledge} are powerful
graph structures that provide structured access mechanisms to knowledge and lie at the heart of key applications like structured or semantic search engines, virtual assistants, and question answering systems. One widely popular representation of knowledge graphs is to use dense continuous representations of entities and relations by embedding them in latent continuous vector spaces. Consequently several knowledge graph embedding methods (e.g., \textsc{RESCAL}~\cite{nickel2011three}, \textsc{TransE}~\cite{bordes2013translating}, \textsc{DistMult}~\cite{yang2014embedding}, \textsc{TransD}~\cite{ji2015knowledge}, \textsc{ComplEx}~\cite{trouillon2016complex}, \textsc{ConvE}~\cite{dettmers2017convolutional}) have been proposed over recent years. 

\begin{figure*}[t!]
\centering
\begin{subfigure}{0.48\textwidth}
	\frame{\includegraphics[width=\textwidth]{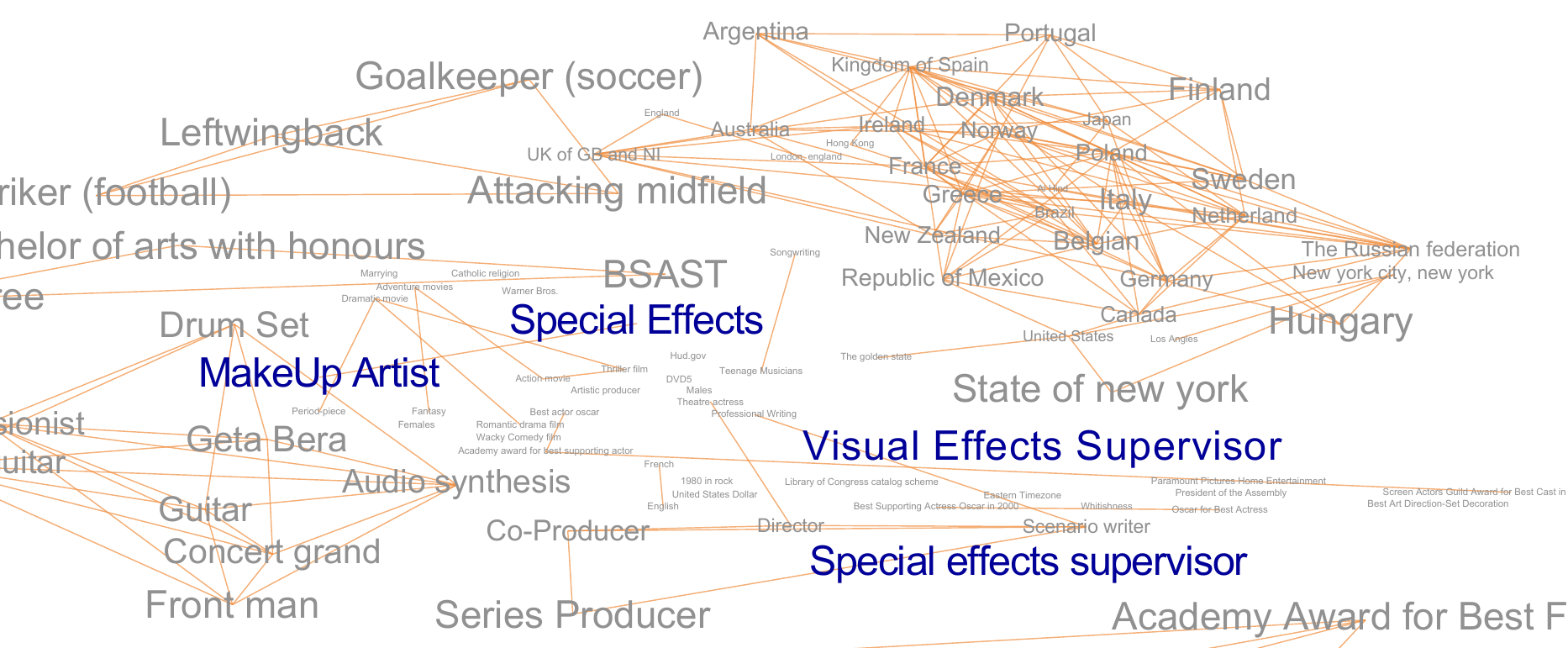}}
	\caption{\textsc{TransE}}
	\label{fig:transe}
\end{subfigure}\hfill
\begin{subfigure}{0.48\textwidth}
\centering
	\frame{\includegraphics[width=\textwidth]{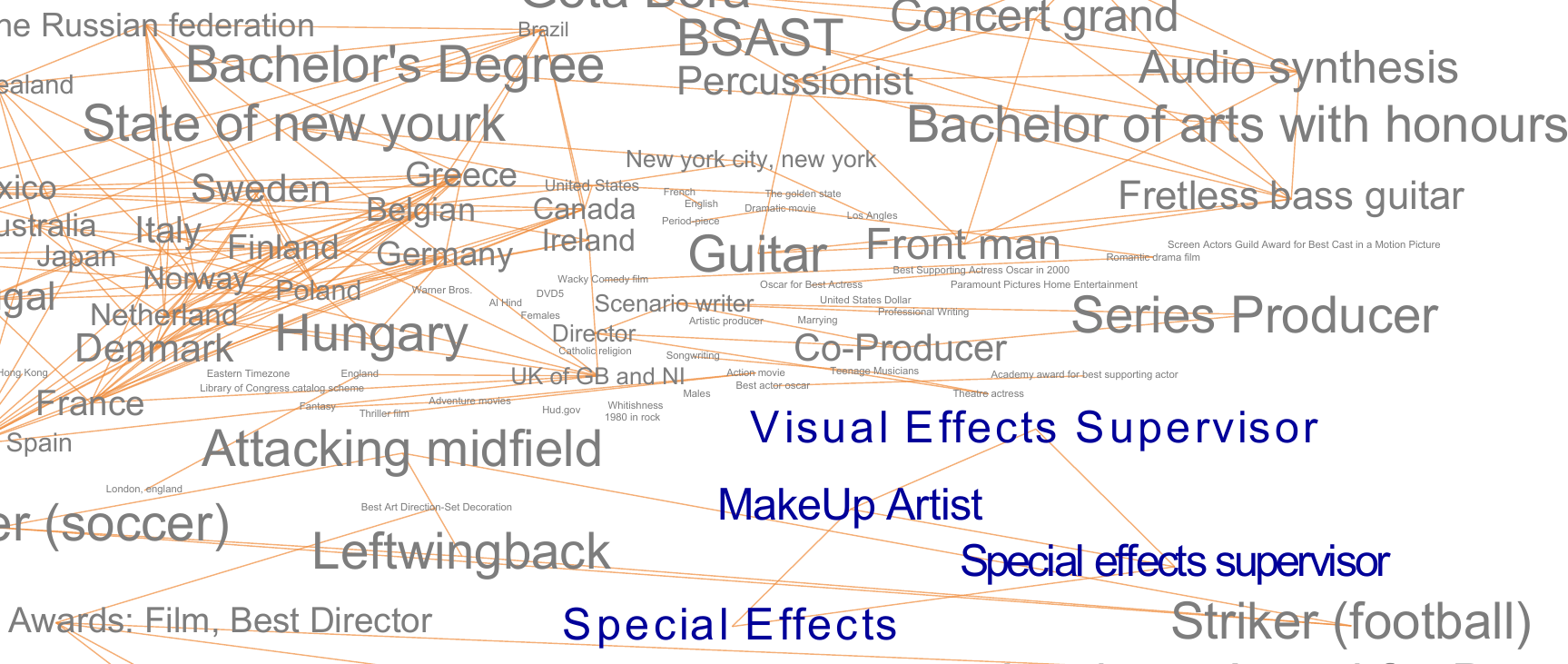}}
	\caption{\textsc{\ourmethod}}
	\label{fig:mohone}
\end{subfigure}
\caption{Note the effect of including higher order (local and global) network cues into Knowledge Graph embeddings.  In the baseline \textsc{TransE} embeddings (Figure \ref{fig:transe}), the entities \texttt{Special Effects, MakeUp Artist} are further away from \texttt{Visual Effects Supervisor, Special Effects Supervisor}. However, since in the knowledge graph these entities share local neighborhoods, \textsc{\ourmethod} captures this latent structure resulting in improved embeddings (Figure \ref{fig:mohone}) where these entities are closer (best viewed in color).}
\label{fig:crown_jewel}
\end{figure*}

Most of these knowledge graph embedding methods typically operate on triples and learn dense continuous representations either using simple shallow linear models or neural models with convolutional layers. Because these methods operate only on triples, such methods are limited by a fairly \textbf{local} view of the entire knowledge graphs and thus are unable to better leverage cues both local and global from the entire knowledge graph. Consequently, in this work, we propose a new framework \textsc{\ourmethod} to effectively leverage network cues from the knowledge graphs to construct ``network-infused'' knowledge graph embeddings. Figure \ref{fig:crown_jewel} illustrates the effect of incorporating higher order (local and global) network cues into knowledge graph embeddings. Note that in the baseline \textsc{TransE} embeddings (see Figure \ref{fig:transe}), the entities \texttt{Special Effects, MakeUp Artist} are further away from \texttt{Visual Effects Supervisor, Special Effects Supervisor}. However, since in the knowledge graph these entities share common local neighborhoods, our method captures this latent structure resulting in improved ``network-infused'' embeddings shown in Figure \ref{fig:mohone} where these entities are much closer together. 

\textsc{\ourmethod} draws on advances in network representation learning to learn effective representations of entities in the knowledge graph that capture their network properties. In particular, \textsc{\ourmethod} can capture \emph{scale-dependent} shared local neighborhood similarities of entities as well as structural similarities where entities serve similar functions or roles in the underlying network. \textsc{\ourmethod} then incorporates these learned network representations of entities into existing knowledge graph embedding models using a fast, efficient iterative updating mechanism. Since our framework is agnostic of the specific knowledge graph embedding method, it can be applied to improve the performance of many existing knowledge graph embedding methods.   

In a nut-shell our contributions are:
\begin{enumerate}
\item We propose a new framework \textsc{\ourmethod}, that effectively incorporates network cues spanning the entire knowledge graphs to learn richer knowledge graph embeddings.
\item Within this framework, we propose a unified learning approach to learning network embeddings that explicitly model the network at a particular scale. In particular, we propose methods to capture two distinct aspects of network similarity: local shared neighborhood and structural role similarity at multiple scales.
\item We propose a fast, efficient iterative update mechanism to incorporate cues captured by our learned network embeddings into existing knowledge graph embedding methods and thus significantly improving their performance sometimes yielding a relative improvement as large as $\textbf{17\%}$.
\end{enumerate}

\section{Related Work}
\paragraph{Knowledge Graph Embeddings}
There has been a long line of work in knowledge graph embeddings, an excellent survey of which is given by \cite{wang2017knowledge}. \textsc{RESCAL} \cite{nickel2011three} proposed a matrix factorization based approach using a bi-linear scoring function. This line of work was followed up by \cite{bordes2013translating} who introduced \textsc{TransE}, the first of many translational models to be proposed. \textsc{TransE} tries to relate the head and tail entity embeddings by modeling the relation as a translational vector. Building on this line of work, a series of models have been proposed \cite{wang2014knowledge,lin2015learning,ji2015knowledge,yang2014embedding,trouillon2016complex}, each of which introduces more refined constraints. More recently proposed models like \textsc{ManiFoldE} \cite{xiao2015one} attempt to model knowledge graph embeddings as a manifold while \textsc{HolE} \cite{nickel2011three} attempts to model circular correlation between entities and draws inspiration from associative memories. Recently models based on convolutional neural networks have been proposed like \cite{dettmers2017convolutional} which utilized 2D convolutions on input triples using convolutional networks to learn knowledge graph embeddings. However, all of these methods operate only on triples and do not attempt to model the network itself beyond the triple level. Two notable exceptions are the works of \cite{guo2015semantically,palumbo2018knowledge}. First, Guo et al. (\citeyear{guo2015semantically}) uses Laplacian graph regularizers to encode ontological category constraints while Palumbo et al. (\citeyear{palumbo2018knowledge}) use \textsc{node2vec} for item recommendation in knowledge graphs. Our work differs from all of these works. First, we depart from the usual paradigm of focusing only on triples in knowledge graph, but learn network representations that encode \emph{higher order} network connectivity patterns accounting for network scale. Second, in contrast to \cite{guo2015semantically}, we do not require entities to be mapped to specific discrete categories. Finally, differing from \cite{palumbo2018knowledge}, we propose methods to model both shared neighborhood similarity and structural similarity of entities in knowledge graphs in a unified framework. Furthermore, our proposed method is agnostic of specific knowledge graph embedding methods and thus can be easily incorporated in existing models to improve the performances of these models.

\paragraph{Network Embeddings} Recently there has been a surge of work in learning dense representations or embeddings of nodes in a social network to capture similarity of nodes. While classical dimensionality reduction  techniques like Laplacian Eigenmaps \cite{belkin2002laplacian} have been used to embed nodes of a network, recent approaches inspired from word embedding models like Skipgram have been shown to be very effective. Such models are based on random walks and attempt to model the probability of node co-occurrences in truncated random walks. Popular models like \textsc{DeepWalk} \cite{perozzi2014deepwalk} and \textsc{Node2Vec}~\cite{grover2016node2vec} belong to this class of models. Recently Donnat et al.（\citeyear{donnat2018learning}） propose a method based on diffusion wavelets to learn structurally similar node embeddings. Neural models to learn network embeddings have also been developed in recent years \cite{Zhu:2018:DVN:3219819.3220052,Tu:2018:DRN:3219819.3220068,Gao:2018:LLG:3219819.3219947}. Finally, we refer to an excellent survey on network representation learning by \cite{hamilton2017representation}.

While our work undoubtedly builds on the principles of these models, we further propose methods to explicitly model ``scale-specific'' network connectivity patterns for both aspects of network similarity (shared neighborhood and structural) in a unified framework based on heat kernels to learn ``network-infused'' knowledge graph embeddings.

\section{Models and Methods}
\subsection{Problem Formulation}
Given a knowledge graph defined by $G=(E, R)$ where $E$ denotes the set of entities and $R$ denotes the set of relations among those entities, we seek to learn $d$-dimensional embeddings of the entities and relations which are compatible with observed facts and are useful for Knowledge graph prediction tasks like link prediction and triple classification. While previous methods \cite{bordes2013translating,wang2014knowledge,ji2015knowledge,lin2015learning,trouillon2016complex} to embed Knowledge Graphs typically model explicit relations between entities by triples of the form $(s, t, r)$ essentially capturing first order network effects,  our method \ourmethod\ recognizes and additionally models higher order network effects to learn Knowledge Graph embeddings that are infused with higher order network information.

\subsection{Overview of \ourmethod}
\label{sec:overview}
\ourmethod\ consists of two main components:
\begin{enumerate}
\item \textbf{Learning Higher Order Network Embeddings} We model higher order network effects in the given knowledge graph by learning a dense representation of each entity based on the underlying network structure.  Specifically, we learn a mapping function $\bm{F}:E \mapsto \mathcal{R}^{d}$. Additionally,  we observe that a node $u$ may be similar to a node $v$ in two different aspects: 
\begin{itemize}
\item \textbf{Shared Neighborhood}: Here, the nodes $u$ and $v$ that share a common local neighborhood (and thus have common neighbors) have similar embeddings (as shown in Fig.\ref{fig:shared}). 
\item \textbf{Similar Structural Roles}: Here, nodes $u$ and $v$ which have similar structural roles have similar embeddings (as shown in Fig.\ref{fig:struct}). 
\end{itemize}

\begin{figure}[htb!]
\centering
\begin{subfigure}{0.23\textwidth}
	\includegraphics[width=\textwidth]{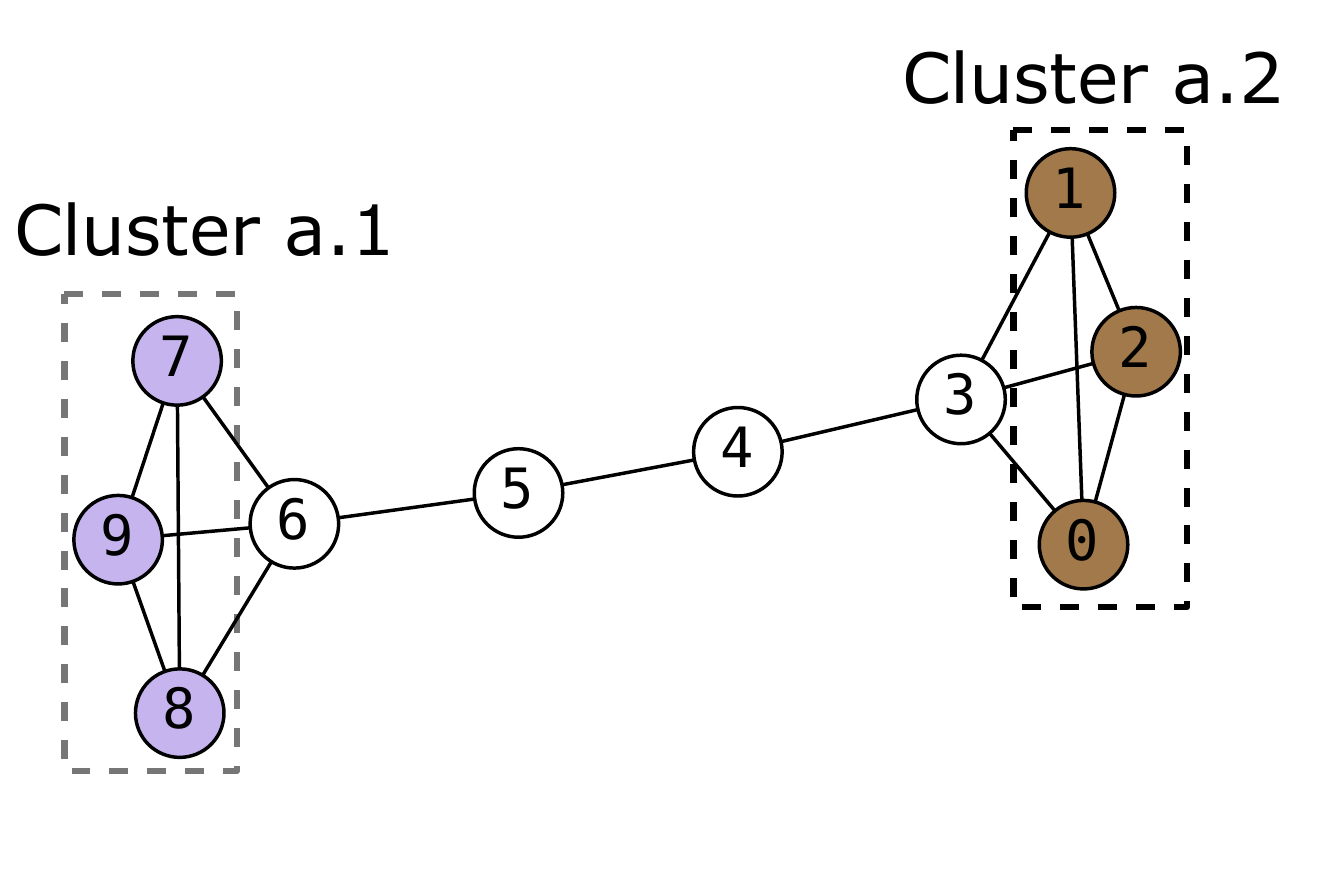}
	\caption{Shared Local Neighborhood}
	\label{fig:shared}
\end{subfigure}
\begin{subfigure}{0.23\textwidth}
\centering
	\includegraphics[width=\textwidth]{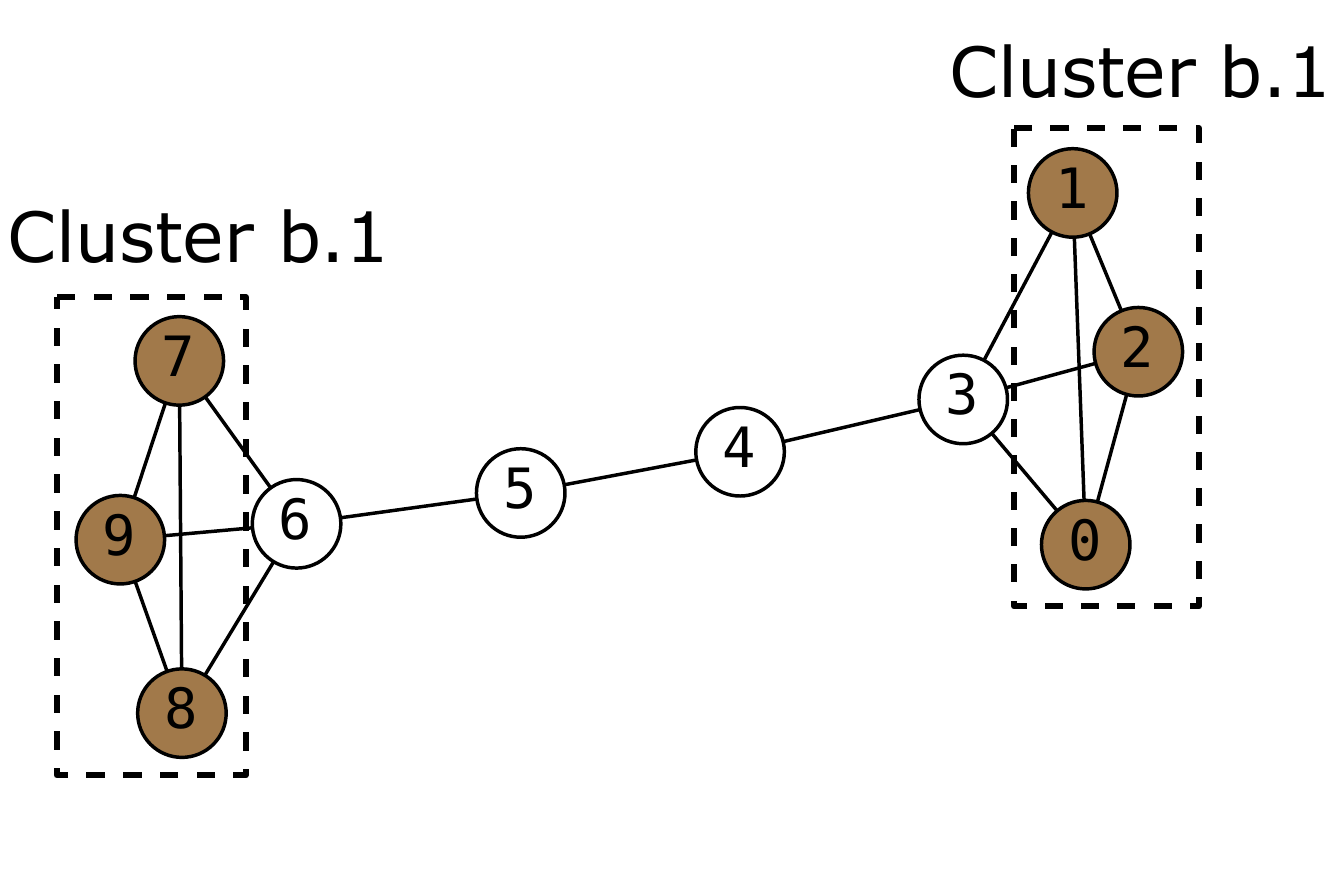}
	\caption{Structural Roles}
	\label{fig:struct}
\end{subfigure}
\caption{\textbf{A schematic network illustrating the two distinct aspects of node similarity in networks.} According to the notion of shared local neighborhood similarity, nodes which share common neighbors are considered similar. In Figure \ref{fig:shared} nodes in the set $\{7,8,9\}$ are similar to each other and nodes in the set $\{0, 1, 2\}$ are similar but nodes belonging to the first group are not similar to nodes in the second group because they are in two different parts of the network and share no local neighbors. On the other hand, according to structural similarity nodes $\{0,1,2,7,8,9\}$ are all similar as they have similar structural roles in the network.}
\label{fig:local_vs_structural}
\end{figure}
Figure \ref{fig:local_vs_structural} illustrates these differences on a toy network. Consequently, we propose methods to model both aspects and learn network embeddings that encode higher order network structure and similarity.
\item \textbf{Infusing Knowledge Graph Embeddings with Higher Order Network Embeddings}: Having learned network embeddings from the previous step, we now propose a simple, fast method that can incorporate the network information encoded in these embeddings into any existing knowledge graph embedding to yield ``network-infused'' knowledge graph embeddings. We first introduce a similarity metric over the learned network embeddings which enables us to quantify the similarity of any two nodes. We then learn a new set of embeddings that seeks to preserve the similarity encoded by the base knowledge graph embeddings while also respecting constraints imposed by network embeddings. In particular, we formulate these constraints as a convex optimization problem which yields an efficient iterative update algorithm.  
\end{enumerate}
We now discuss each of the above components in detail. 

\subsection{Learning Higher Order Network Embeddings}
Here, we describe in detail our methods for learning higher order network embeddings. As noted in the overview of \ourmethod, we model two distinct aspects of network similarity: \textbf{(a) Shared Neighborhood Similarity} and \textbf{(b) Structural Role Similarity}. When modeling both of these aspects, we explicitly model the network at a particular resolution or order (determined by a hyper-parameter) which enables us to encode network similarity patterns based on the full spectrum: from fine-grained local connectivity patterns to coarse global connectivity patterns. We treated the knowledge graph as a plain undirected graph when learning the network embedding. We make a simplifying assumption that modeling the aggregate is still useful in capturing network properties and useful for KG prediction tasks (and leave the more refined case to future work).

We achieve this multi-resolution modeling by noting connections between \emph{diffusion heat matrices over graphs} and \emph{random walks over graphs}. Based on these connections, we further cast the problem of modeling both aspects of network similarity within a unified framework. We now describe our unified framework followed by relevant background on diffusion heat kernels. We then discuss at length our proposed methods to model shared neighborhood similarity and structural role similarity within this unified framework.

\paragraph{Framework for learning network embeddings} Our framework for learning network embeddings is grounded on maximizing the log likelihood of observing the set of nodes $\mathcal{N}(u)$, given the node $u$. Furthermore, each node $v\in \mathcal{N}(u)$ can be associated with a weight indicating a notion of some association between $u$ and $v$. As we will see later, different definitions of $\mathcal{N}(u)$ can be used to capture different aspects of network similarity. We now parametrize the model as follows: 
\begin{enumerate}
\item Let $\bm{F}$ be an embedding matrix of size $(|E|,d)$ where $\bm{F}(u)$ represents the embedding of node $u$, $|E|$ is the total number of entities, and $d$ the embedding dimension.  
\item We then maximize the following likelihood function $\max_{\bm{F}} \sum_{u} \log \Pr(\mathcal{N}(u)|u)$. We also make the simplifying assumption by assuming that the likelihood of observing a node $v\in \mathcal{N}(u)$ is conditionally independent of any other node in the neighborhood given $u$. That is:
\begin{equation}
\log \Pr(\mathcal{N}(u)|u)=\sum_{v\in \mathcal{N}(u)}\log\ \Pr(v|u). 
\end{equation}
\item We then model $\Pr(v|u)=\frac{e^{\bm{F}(u).\bm{F}(v)}}{\sum_{v}e^{\bm{F}(u).\bm{F}(v)}}$. Having fully specified the log likelihood function, we can now optimize it using stochastic gradient ascent and train the model in an online fashion by sampling node pairs $(u, v\in \mathcal{N}(u))$ where $v\neq u$ is sampled according to it's weight (which can be uniform by default). 
\end{enumerate}

Finally, in order to learn network embeddings within this framework, we only need to specify $\mathcal{N}(u)$ and the corresponding weights associated with each node in $\mathcal{N}(u)$. 

\paragraph{Diffusion Heat Matrices on Graphs}
Let $\mathbf{A}$ be the adjacency matrix of a graph, and $\mathbf{D}$ is the corresponding degree matrix. The normalized graph Laplacian matrix is defined as $\mathbf{L}=\mathbf{I}-\mathbf{D}^{-\frac{1}{2}}\mathbf{AD}^{-\frac{1}{2}}$. Consider the eigen decomposition of $\mathbf{L}=\mathbf{U\Lambda U^T}$.  $\mathbf{\Lambda} = \text{Diag}(\lambda_1,\lambda_2,\dots,\lambda_N))$ is the diagonal matrix of the eigenvalues in ascending order and $\mathbf{U}$ are the corresponding eigen vectors arranged in columns. 
Define $g_s=e^{-\lambda s}$ as a heat kernel with scale $s$. The heat diffusion pattern matrix is defined as:
\begin{equation}
\mathbf{\Psi} = \mathbf{U}\text{Diag}(g_s(\lambda_1),\dots,g_s(\lambda_N))\mathbf{U^T}
\end{equation}
We note the following properties of $\mathbf{\Psi}$:
\begin{enumerate}
\item The scale $s$ controls the degree to which local connectivity patterns are captured. Small values of $s$ tend to capture more local connectivity patterns while larger values of $s$ capture more global patterns. In particular, when $s$ tends to $0$, then $\mathbf{\Psi} =\mathbf{I}-\mathbf{L}s$ suggesting it depends only on the local connectivity structure or the topology of the graph.  In contrast, when $s$ is large, $\mathbf{\Psi} = e^{-s\lambda_{min}}\mathbf{U_{min}}\mathbf{U_{min}^T}$, where $\lambda_{min}$ is the smallest non-zero eigenvalue and $\mathbf{U_{min}}$ is the corresponding eigenvector. 
\item Each column $\Psi_{a}$ can be interpreted as a distribution of heat over all the nodes of the graph. In particular it can be interpreted as the amount of heat at each node after time $t=s$, assuming a unit heat source placed at node $a$.
\end{enumerate}
Thus the diffusion matrix $\mathbf{\Psi}$ models network connectivity patterns at a specific resolution with the additional property that it is column stochastic. We now discuss how we use $\mathbf{\Psi}$ to learn network embeddings according to our framework. 

\subsubsection{Modeling Shared Neighborhood: \textsc{\ourmethod-shnb}}
To capture the aspect of shared neighborhood similarity, we learn network embeddings using the unified framework described. Recall that we only need to specify $\mathcal{N}(u)$ and the weight associated with each node in $\mathcal{N}(u)$. In particular, we set $\mathcal{N}(u)$ to be the set of all nodes. The weight associated with a node $v\in \mathcal{N}(u)$ is given by $\mathbf{\Psi}_{u}(v)$. Note that since $\mathbf{\Psi}$ is a column stochastic matrix, $\mathbf{\Psi}_{u}$ defines a probability distribution over the node set of the network and thus defines the weights associated for each node in $\mathcal{N}(u)$.
We will denote embeddings learned using this specification as \textsc{\ourmethod\ -shnb}.

\emph{Connections to Random Walk based Network Embedding models} Recall that popular network embedding models like Node2Vec \cite{grover2016node2vec} or DeepWalk \cite{perozzi2014deepwalk} use truncated random walks to capture node co-occurrences. At their heart, both of these methods model pair-wise node co-occurrences exactly as described in our framework. In fact, as noted by \cite{hamilton2017representation}, they can essentially be viewed as implicitly sampling node pairs from a truncated random walk transition matrix. In particular, given node $u$ the node $v$ is implicitly sampled from a distribution $\pi_{u}$ where $\pi_{u}(v)$ specifies the probability of visiting $v$ on a length-T random walk starting at $u$.

Our proposed model which in-turn samples from $\mathbf{\Psi}$ can be similarly interpreted in terms of random walks. Specifically, $\mathbf{\Psi}_{u}(v)$ can be interpreted as the probability of a \emph{lazy random walk}\footnote{Recall that a lazy random walk is similar to a random walk but has a probability of $\beta$ of staying at the current node.} starting at node $u$ and ending at node $v$ where a parameter $\beta$ which specifies the probability of just staying at the current node is a function of $s$, the scale parameter of the heat kernel. In other words,  our method can be viewed as a generalization of \textsc{Node2Vec} which can explicitly model network orders or scales (as determined by the heat kernel) and can be viewed as implicitly using lazy random walks.

\begin{table*}
\small
\centering
	\begin{tabular}{c|c|c|c}
    \hline
    \textbf{Method} & \textbf{Node Pairs Sampling Distribution} & \textbf{Random Walk View}  & \textbf{Explicitly model network order}\\
    \hline
    \textsc{DeepWalk} & $\pi_{u}^{{trw}}$ & Truncated Random Walks & No \\
    \textsc{Node2Vec} & $\pi_{u}^{{btrw}}$ & Biased Truncated Random Walks & No\\
    \textsc{\ourmethod-shnb} & $\mathbf{\Psi}_{u}$ & Lazy Random Walks & Yes (Heat Kernel scale: $s$) \\ 
     \textsc{\ourmethod-struct} & -- & -- & Yes (Heat Kernel scale: $s$)  \\ \hline 
   \end{tabular}
    \caption{\small{A quick summary of how popular network embedding approaches like \textsc{DeepWalk} \cite{perozzi2014deepwalk} and \textsc{Node2vec} \cite{grover2016node2vec} as well as \textsc{\ourmethod-shnb} can all be placed within our unified framework.} Given a node $u$, \textsc{DeepWalk} implicitly samples the node $v$ from a length-T truncated random walk transition matrix denoted by $\pi_{u}^{{trw}}$ which specifies the probability of visiting $v$ on a length-T random walk starting at $u$. \textsc{Node2Vec} is similar to \textsc{DeepWalk} but uses a transition matrix resulting from biased random walks denoted by $\pi_{u}^{{btrw}}$.
\textsc{\ourmethod-shnb}, our method can be viewed as a generalization of \textsc{Node2Vec} which can explicitly model network orders or scales (as determined by the heat kernel) and can be viewed as implicitly using lazy random walks. 
   }
\end{table*}

\subsubsection{Modeling Structural Roles: \textsc{\ourmethod-struct}}
\begin{figure}[t!]
\centering
\begin{subfigure}{0.23\textwidth}
	\includegraphics[width=\textwidth]{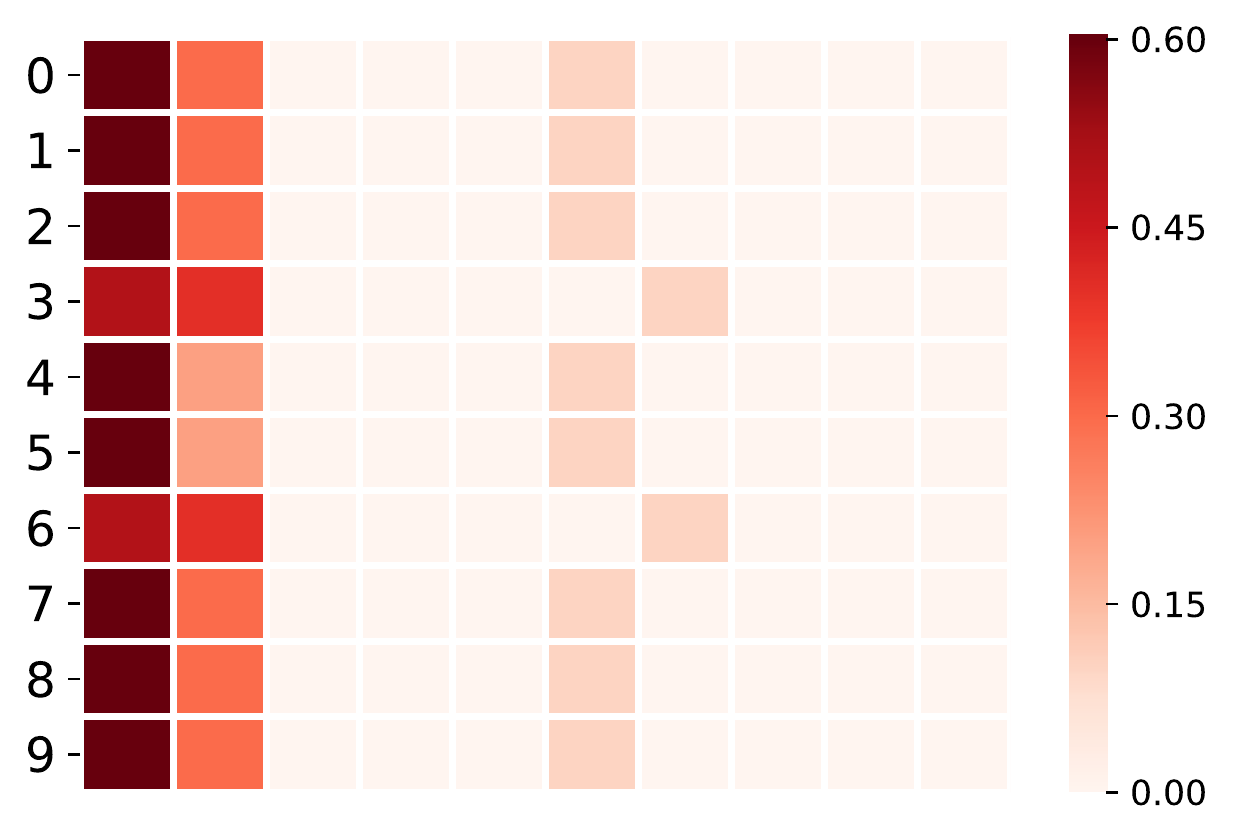}
	\caption{Empirical Distribution of heat values for each node.}
	\label{fig:struct_heatmap}
\end{subfigure}
\begin{subfigure}{0.23\textwidth}
\centering
	\includegraphics[width=\textwidth]{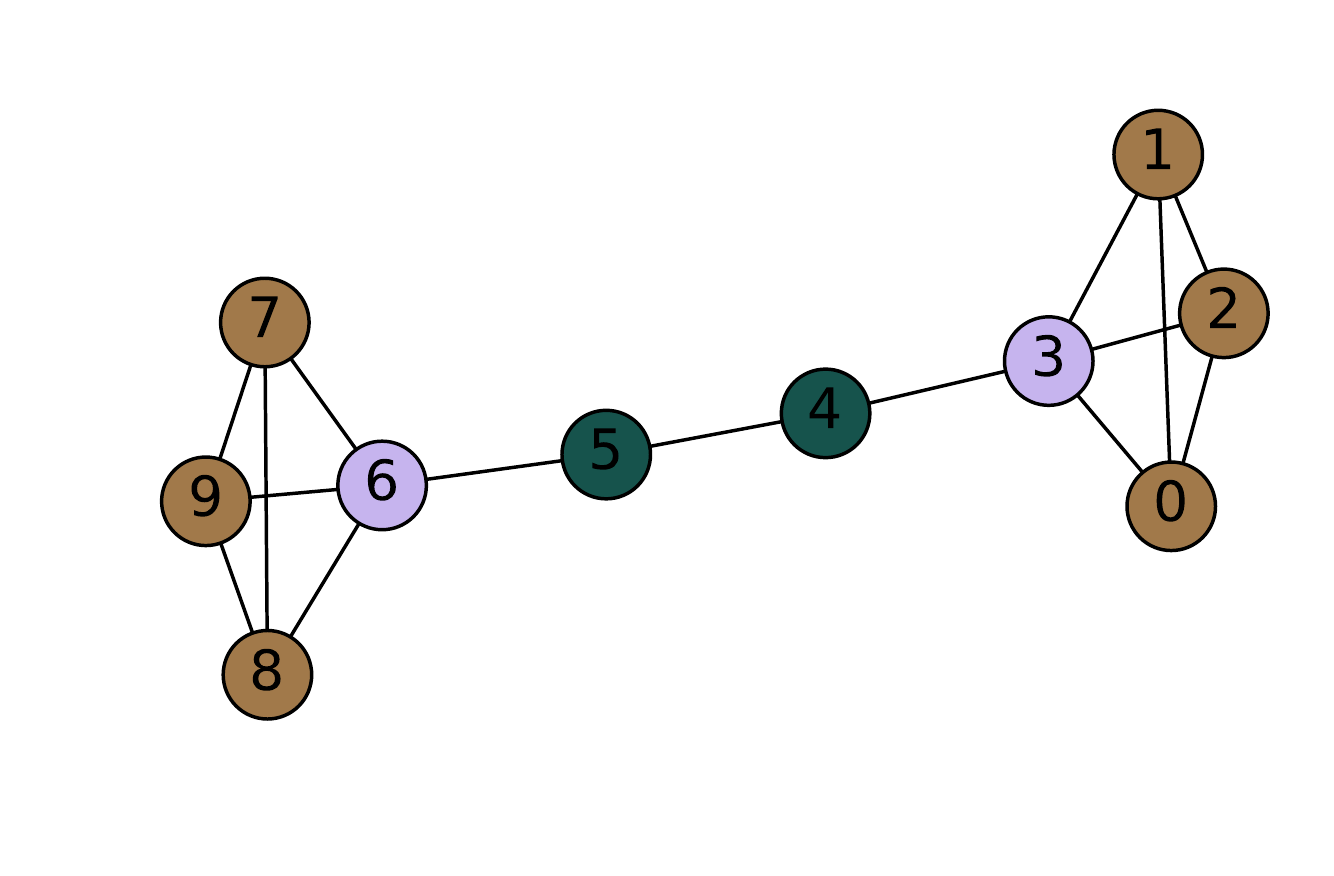}
	\caption{Identified node clusters based on Structural similarity.}
	\label{fig:network}
\end{subfigure}
\caption{Distribution of values for $\Psi_{v}$ visualized as a heat map for each node in the bar-bell graph shown in Figure \ref{fig:network}. Note that nodes with similar heat patterns automatically reveal $3$ clusters of structurally similar nodes (Best viewed in color).}
\label{fig:struct_insight}
\end{figure}

In the previous section, we outlined how we used the diffusion pattern matrix $\mathbf{\Psi}$ within our unified framework to learn network embeddings that captured local neighborhood similarity. In this section, we now turn our attention to modeling the second aspect of network similarity namely structural role similarity. In particular, we would like two nodes $u$ and $v$ to have similar embeddings if they have similar structural roles in the network. To achieve this, we rely on the following key observation about the columns of $\mathbf{\Psi}$ as noted by \cite{donnat2018learning}: \emph{When two nodes (u, v) are structurally similar, the distributions of heat kernel values in $\mathbf{\Psi}_{u}$ and $\mathbf{\Psi}_{v}$ are similar} (see Figure \ref{fig:struct_insight} for an illustration.)

This immediately enables us to model structural similarity in our framework as follows:
\begin{enumerate}
\item Define $D(u, v)$ to be the Jensen-Shannon divergence between the distributions of values in $\mathbf{\Psi}_{u}$ and $\mathbf{\Psi}_{v}$. 
\item For a given node $u$, define $\mathcal{N}(u)$ as the set of all nodes. 
\item The weight associated with node $v\in\mathcal{N}(u)$ is given by standardizing $D(u,v)$ across all nodes in $\mathcal{N}(u)$. 
\end{enumerate}

Intuitively, since $\mathcal{N}(u)$ approximates nodes which are structurally similar to $u$, our unified embedding framework will seek to learn similar embeddings for $u$ and $v\in\mathcal{N}(u)$. Furthermore, since the heat kernel captures connectivity patterns at scale $s$ in $\mathbf{\Psi}$, our learned embeddings in turn explicitly model connectivity at that scale (or order).  

\subsection{Infusing Knowledge Graph Embeddings with Higher Order Network Embeddings}
Having learned higher order network embeddings for all entities in the knowledge graph, we now describe the second main component of \textsc{\ourmethod} -- infusing knowledge graph embeddings with the learned network embeddings $\bm{F}$.

Specifically, let $\hat{\mathbf{K}}=(\hat{\mathbf{Q}}, \hat{\mathbf{W}})$ denote a set of Knowledge graph embeddings learned by any Knowledge Graph Embedding method (like TransE, TransD, RESCAL etc.) where $\hat{\mathbf{Q}}$ denotes the set of entity embeddings and $\hat{\mathbf{W}}$ denotes the set of relation embeddings. 

We seek to learn a new knowledge graph embedding $\mathbf{K}=(\mathbf{Q}, \mathbf{W})$ which also encodes the structure encoded by the learned network embeddings $\bm{F}$. We operationalize this by mathematically modeling the following constraints:
\begin{enumerate}
\item \textbf{Prior Constraint}: Seek to preserve similarity of new entity embeddings $\mathbf{Q}$ with the given entity embeddings $\hat{\mathbf{Q}}$. We model this constraint by seeking to learn $\mathbf{Q=(q_{1}, q_{2}...q_{|E|})}$ such that its columns are close in Euclidean distance to their counterparts in $\hat{\mathbf{Q}}$.  
\item \textbf{Network Constraint}: Incorporate the structure encoded by network embeddings $\bm{F}$ into the new entity embeddings $\mathbf{Q}$. To do this, we seek to learn $\mathbf{Q}$ such that the embedding for node $i$ namely $\mathbf{q_{i}}$ is also close in Euclidean distance to the nodes in the set $\Omega_{i}$, the set of nearest neighbors of $i$ in the network embedding space $\bm{F}$.  
\end{enumerate}

Formally, the above constraints can be encoded in the following minimization problem over the parameters defined by $\mathbf{Q}$: 
\begin{equation}
	\Theta(\mathbf{Q})=\sum^{|E|}_{i=1}\left[{\alpha_i {\Vert \mathbf{q_{i}} - \hat{\mathbf{{q}_i}} \Vert}^2 + \sum_{(i,j)\in \Omega_{i}}{\beta_{ij}{\Vert \mathbf{{q}_i} - \mathbf{{q}_j} \Vert}^2}}\right]
	\label{equ_obj_retro}
\end{equation}, where the set of hyper-parameters $\alpha$ and $\beta$ controls the relative strengths of these constraints.

The above optimization problem can be interpreted as a Markov random field~\cite{doi:10.1080/0022250X.1980.9989895}, where nodes $\hat{\mathbf{q_i}}$ are observed while nodes $\mathbf{q_i}$ are to be estimated. Moreover, the above optimization problem is convex in $\mathbf{Q}$ and therefore can be solved very efficiently using iterative updates \cite{2014arXiv1411.4166F,bengio200611,subramanya2010efficient,das2011unsupervised,das2011semi}. In particular, the iterative update algorithm involves the following:
\begin{itemize}
\item Initialize $\mathbf{Q}$ with $\hat{\mathbf{Q}}$ and apply the following online iterative updates.
\item Update the parameter $\mathbf{q_{i}}$ using the following equation:
\begin{equation}
	\mathbf{q_i}=\frac{\sum_{j:(i,j)\in \Omega_{i}}\beta_{ij}\mathbf{q_j}+\alpha_i\hat{\mathbf{q_i}}}{\sum_{j:(i,j)\in \Omega_{i}Z}\beta_{ij}+\alpha_i}
	\label{equ_itr_retro}
\end{equation}
The update for parameter $\mathbf{q_i}$ is derived by computing the derivative of the objective $\Theta(\mathbf{Q})$ with respect to $\mathbf{q_i}$ and equating it to $0$. 
\end{itemize}

The above iterative updates solution generally converges after 10 iterations, and is fast in practice. Furthermore, the updating process is model agnostic, and thus can be applied to arbitrary knowledge graph embedding methods.

Finally, having learned network infused entity embeddings $\mathbf{Q}$, we can easily learn the corresponding relation embeddings $\mathbf{W}$ by simply initializing the chosen knowledge graph embedding method (which was used to derive $\hat{\mathbf{K}}=(\hat{\mathbf{Q}}, \hat{\mathbf{W}})$) with the newly learned network infused entity embeddings $\mathbf{Q}$ and learning the relation embeddings as dictated by the method. 

\paragraph{Scalability and Complexity of \ourmethod} \ourmethod\ easily scales to large data-sets. Computing $\mathbf{\Psi}$ requires computing powers of the Laplacian. Each power can be computed in $\mathcal{O}(|\mathcal{E}|)$ time (linear in the number of edges:$|\mathcal{E}|$). Specifically, $\mathbf{\Psi}$ can be computed using an approximation relying on $K$-degree Chebychev polynomials \cite{donnat2018learning} requiring  $\mathcal{O}(K|\mathcal{E}|)$ time . Moreover, the rest of the training procedure has the same computational efficiency as the highly scalable \textsc{Node2Vec} which is amenable to easy parallelization and online training. Finally, the iterative update mechanisms to learn ``infused'' knowledge graph embeddings is very fast (involving only simple additive and multiplicative updates) and converges in only $10$ iterations.  

\section{Experiments}
Here we evaluate our proposed methods on the task of link prediction. We emphasize that our method is agnostic of the actual knowledge graph embedding method used. To illustrate this, we consider a battery of well-established baselines and evaluate the improvement derived on applying our method on top of each baseline considered.

In particular, we evaluate all methods on two standard data-sets: FB15K237~\cite{toutanova2015observed} and WN18RR~\cite{dettmers2017convolutional}\footnote{Since we obtain similar observations on both datasets we report performance in WN18RR in the supplementary material.} which are used extensively to evaluate link prediction in Knowledge graphs \footnote{Since these datasets are standard, we refer the reader to prior works like \cite{toutanova2015observed}, and \cite{dettmers2017convolutional} for statistics regarding the datasets.}. 

As baselines, we consider several state-of-the-art knowledge graph embedding methods. These methods can be roughly categorized into two groups~\cite{wang2017knowledge}: translational distance models (which adopt a distance-based scoring function) and semantic matching models (which adopt a similarity-based scoring function). The translational distance models include \textsc{TransE}~\cite{bordes2013translating}, \textsc{TransH}~\cite{wang2014knowledge}, \textsc{TransR}~\cite{lin2015learning}, \textsc{TransD}~\cite{ji2015knowledge} while the semantic matching models we consider are 
\textsc{RESCAL}~\cite{nickel2011three}, \textsc{DistMult}~\cite{yang2014embedding}, \textsc{ComplEx}~\cite{trouillon2016complex}, \textsc{Analogy}~\cite{liu2017analogical}.

In-line with prior work \cite{bordes2013translating}, we report the standard evaluation metrics: the average of the reciprocal rank of all correct entities (Mean Reciprocal Rank), the proportion of correct entities ranked in top10 (Hits@10) in the filtered and type-constrained setting exactly as described by \cite{bordes2013translating,toutanova2015observed}\footnote{We use the freely available OpenKE code to compute this.}. 

\subsection{Experimental Settings}
We trained all baseline models and obtained baseline knowledge graph embeddings using \textsc{OpenKE}\footnote{The code is available here: http://openke.thunlp.org/}. The batch-size was set to $100$ with early-stopping while all other hyper-parameters such as the learning rate, or the optimization algorithm (SGD, Adam, AdaGrad) were selected using a grid-search on the validation set.  To obtain \textsc{\ourmethod-shnb} embeddings and \textsc{\ourmethod-struct}, we set the scale parameter $s=5$, the embedding dimension $d=100$ and train until convergence after training for a minimum of $10$ epochs. 

\subsection{Results and Analysis}
\label{sec:results}

\begin{table*}[htb!]
\small
\centering
\begin{subtable}{1.0\textwidth}
	\centering
	\begin{tabular}{|c|c|cc|cc|cc|}
    \textbf{\textsc{MRR}} & \textbf{\textsc{Baseline}} & \multicolumn{2}{c|}{\textbf{\textsc{Baseline+\ourmethod}}} & \multicolumn{2}{c|}{\textbf{\textsc{Increase (Absolute)}}} & \multicolumn{2}{c|}{\textbf{\textsc{Increase (Relative)}}} \\
    \hline
    & & \textbf{\textsc{SHMB}} & \textbf{\textsc{Struct}} & \textbf{\textsc{SHMB}} & \textbf{\textsc{Struct}} & \multicolumn{1}{c}{\textbf{\textsc{SHMB}}} & \multicolumn{1}{c|}{\textbf{\textsc{Struct}} }\\
    \textsc{TransE} & 29.6 & $31.5^{\star}$ & $30.3^{\star}$ & +1.9 & +0.7 & +6.4\% & +2.3\% \\
    \textsc{TransH} & 24.2 & $28.5^{\star}$ & 24.4 & +4.3 & +0.2 & +17.8\% & +0.8\% \\
    \textsc{TransR} & 28.0 & $28.7^{\star}$ & 27.4 & +0.7 & -0.6 & +2.5\% & -2.1\% \\
    \textsc{TransD} & 25.1 & $29.0^{\star}$ & 25.5 & +3.9 & +0.4 & +15.5\% & +1.6\% \\
    \textsc{RESCAL} & 26.8 & $29.2^{\star}$ & $27.9^{\star}$ & +2.4 & +1.1 & +9.0\% & +4.1\% \\
    \textsc{DistMult} & 26.1 & $28.5^{\star}$ & 26.4 & +2.4 & +0.3 & +9.2\% & +1.1\% \\
    \textsc{ComplEx} & 26.1 & $30.0^{\star}$ & $28.4^{\star}$ & +3.9 & +2.3 & +14.9\% & +8.8\%\\
    \textsc{Analogy} & 26.6 & $29.2^{\star}$ & $27.8^{\star}$ & +2.6 & +1.2 & +9.8\% & +4.5\%\\
    \hline
    \end{tabular}
    \caption{Mean Reciprocal Rank on FB15K-237 }
    \label{tab:mrr_fb}
\end{subtable}
\begin{subtable}{1.0\textwidth}
\centering
	\begin{tabular}{|c|c|cc|cc|cc|}
        \textbf{\textsc{Hits@10}} & \textbf{\textsc{Baseline}} & \multicolumn{2}{c|}{\textbf{\textsc{Baseline+\ourmethod}}} & \multicolumn{2}{c|}{\textbf{\textsc{Increase (Absolute)}}} & \multicolumn{2}{c|}{\textbf{\textsc{Increase (Relative)}}}\\
    \hline
     & & \textbf{\textsc{SHMB}} & \textbf{\textsc{Struct}} & \textbf{\textsc{SHMB}} & \textbf{\textsc{Struct}} & \multicolumn{1}{c}{\textbf{\textsc{SHMB}}} & \multicolumn{1}{c|}{\textbf{\textsc{Struct}} }\\
    \textsc{TransE} & 47.3 & $49.4^{\star}$ & $48.0^{\star}$ & +2.1 & +0.7 & +4.4\% & +1.5\% \\
    \textsc{TransH} & 41.7 & $46.4^{\star}$ & 41.3 & +4.7 & -0.4 & +11.2\% & -1.0\% \\
    \textsc{TransR} & 45.2 & $45.5^{\star}$ & 44.2 & +0.3 & -1.0 & +0.7\% & -2.2\% \\
    \textsc{TransD} & 43.0 & $46.8^{\star}$ & 42.3 & +3.8 & -0.7 & +8.8\% & -1.6\% \\
    \textsc{RESCAL} & 43.7 & $46.5^{\star}$ & $45.2^{\star}$ & +2.8 & +1.5 & +6.4\% & +3.4\% \\
    \textsc{DistMult} & 44.9 & 45.0 & 44.5 & +0.1 & -0.4 & +0.2\% & -0.9\% \\
    \textsc{ComplEx} & 45.4 & $46.7^{\star}$ & 44.9 & +1.3 & -0.5 & +2.8\% & -1.1\% \\
    \textsc{Analogy} & 45.5 & $45.9^{\star}$ & 44.7 & +0.4 & -0.8 & +0.9\% & -1.8\% \\
    \hline
    \end{tabular}
    \caption{Hits@10 Performance on FB15K-237}
    \label{tab:hit@10_fb}
\end{subtable}
\caption{{Performance improvements of link prediction on several Knowledge Graph embedding baselines when \ourmethod\ is incorporated. $\star$ indicates improvement over baseline is statistically significant at $0.05$ level.}}
\label{tab:main_results}
\end{table*}

\begin{table*}[htb!]
\small
	\centering
	\begin{tabular}{|c|c|cc|cc|}
    \textsc{\textbf{Model}} & \textbf{\textsc{Baseline}} & \textbf{\textsc{+Node2vec}} & \textbf{\textsc{+\ourmethod-shnb}} & \multicolumn{2}{c|}{\makecell{\textbf{\textsc{Improvement}} \\ \textbf{\textsc{\ourmethod\ vs Node2Vec}}}} \\ \hline
    & & & & \textbf{\textsc{Absolute}} & \textbf{\textsc{Relative}}  \\
    \textsc{TransE} & 29.6 & 31.5 & 31.5 & +0 & +0\% \\
    \textsc{TransH} & 24.2 & 26.9 & $28.5^{\star}$ & +1.6 & +5.94\% \\
    \textsc{TransR} & 28.0 & 28.4 & 28.7 & +0.3 & +1.06\% \\
    \textsc{TransD} & 25.1 & 27.8 & $29.0^{\star}$ & +1.2 & +4.32\% \\
    \textsc{RESCAL} & 26.8 & 28.9 & $29.2^{\star}$ & +0.3 & +1.04\% \\
    \textsc{DistMult} & 26.1 & 27.2 & $28.5^{\star}$ & +1.3 & +4.78\% \\
    \textsc{ComplEx} & 26.1 & 27.9 & $30.0^{\star}$  & +2.1 & +7.53\% \\
    \textsc{Analogy} & 26.6 & 27.6 & $29.2^{\star}$ & +1.6 & +5.80\% \\
    \hline
    \end{tabular}
    \caption{Our network embedding method which explicitly incorporates scale generally outperforms Node2Vec (yielding an average improvement of $3.8\%$ over all models). $\star$ indicates improvement is statistically significant at $0.05$ level.}
     \label{tab_heat_kernel}
\end{table*}

Table \ref{tab:main_results} shows the improvement in terms of Mean Reciprocal Rank (see Table \ref{tab:mrr_fb}) and Hits@10 (see Table \ref{tab:hit@10_fb}) achieved by several models when they incorporate our method \ourmethod. Examining these tables, we make the following observations and conclusions.

\begin{itemize}
\item \textbf{Overall \textsc{\textbf{\ourmethod}} improves existing baseline models}: First, note that incorporating \textsc{\ourmethod} generally significantly improves the performance of many baseline models including \textsc{TransD, TransE, Rescal, Complex, Analogy}. Overall, we note an average improvement across all models of  
$\textbf{2.76}$ points or by $\textbf{10.63\%}$ in terms of MRR (see Table \ref{tab:mrr_fb}). For example, incorporating our method improves the MRR of TransH by $\textbf{4.3}$ points (by $\textbf{17.8\%}$), TransD by $\textbf{3.9}$ points (by $\textbf{15.5\%}$). We also note that the size of the improvement varies for different baseline models with a  number of models like \textsc{TransE, TransH, TransD, RESCAL,  DistMult, ComplEx, Analogy} showing improvements greater than $\textbf{6\%}$ in terms of MRR.  Furthermore, by examining Table \ref{tab:hit@10_fb}, we note that incorporating \ourmethod \ not only improves the overall ranking (in terms of MRR) of existing models, but also improves the hit rate (hits@10) as well. Overall across all baseline models, we note an average relative improvement of $\textbf{5.7\%}$ on Hits@10. Altogether, these observations suggest that \ourmethod\  which incorporates higher order network information across the entire network can significantly improve existing baseline models.

\item \textbf{Shared neighborhood network similarity is better than structural similarity}: \emph{For link prediction, which aspect of network similarity is superior: Shared Neighborhood based similarity or Structural Role based similarity?} By examining Table \ref{tab:mrr_fb} to compare which aspect of network similarity yields a better improvement, we see that shared neighborhood similarity \textsc{\ourmethod-SHNB} is unanimously superior to structural role based similarity \textsc{\ourmethod-Struct} on the task of link prediction. Note that the improvement yielded by \textsc{\ourmethod-Struct} is generally quite small when compared to \textsc{\ourmethod-SHNB} even though there are exceptions like \textsc{ComplEx} where incorporating structural similarity yields very good improvements. This suggests that in the case of link prediction, a broader view of the shared neighborhood is generally more informative than structural roles of nodes. 
\item \textbf{It is better to explicitly incorporate the scale when modeling shared neighborhood similarity} In contrast to existing network representation methods like \textsc{DeepWalk} or \textsc{Node2Vec} \cite{grover2016node2vec}, \ourmethod\ explicitly incorporates a scale parameter ``s'' that influences the degree of network connectivity modeled. Small values of ``s'' help capture local fine-grained connectivity patterns determined by graph topology whereas large values emphasize more global connectivity patterns.

Here, we gauge the importance of explicitly modeling scale by comparing \textsc{\ourmethod-SHNB} to \textsc{Node2Vec} -- a very strong method to learn network embeddings that does not explicitly model the scale. We repeat all our experiments by replacing network embeddings learned using \textsc{\ourmethod-SHNB} with network embeddings learned using \textsc{Node2Vec}. 

Table \ref{tab_heat_kernel} shows the results of this experiment. We note right away, that \textsc{\ourmethod} network embeddings consistently perform at least as well as \textsc{Node2Vec} network embeddings and even significantly out-perform \textsc{Node2Vec}. On an average, across all baseline KG models, \textsc{\ourmethod} network embeddings yield a relative improvement of $\textbf{3.8\%}$ over \textsc{node2Vec} network embeddings, suggesting the importance of explicitly modeling scale and varying degrees of network connectivity.

\end{itemize}

\section{Conclusion}
We proposed \textsc{\ourmethod}, a new framework that incorporates higher order network effect information into existing knowledge graph embedding methods. In particular, we present methods to incorporate different aspects of network similarity including local neighborhood based node similarity and structural role based similarity by learning network representations of entities that capture varying patterns of network connectivity in the Knowledge Graph all within a unified learning framework.  We then develop a fast, efficient iterative update method to incorporate these representations into existing knowledge graph embedding models and demonstrate that our method significantly improves the performance of these models on knowledge graph prediction tasks like link prediction sometimes by at least $\textbf{17\%}$. \ourmethod\ is very efficient and scalable, given that computing the heat kernel is linear in the number of edges and learning the network embeddings is amenable to efficient parallelization and online-training. Our analysis also reveals that for the predictive task of link prediction, modeling shared local neighborhood similarity is more informative than modeling structural role based similarity  empirically validating the differing aspects of network similarity captured.

Our work also suggests a few directions for future research. First, while both structural and share neighborhood similarity empirically yield improvements, how can one explain what fine-grained node measures correlate with such improvements? How do the characteristics of the network (measured in terms of clustering coefficients, modularity) correlate with performance on predictive tasks? A second direction is to investigate a scalable joint training approach to infuse network cues into training procedures for existing knowledge graph embedding methods. Finally, another direction would be to precisely analyze the trade-offs of modeling the different aspects of network similarity on other predictive tasks associated with knowledge graphs.   Altogether, our proposed framework effectively leverages network cues from the entire knowledge graph to significantly improve the performance of several models on predictive tasks on knowledge graphs. 

\bibliography{naaclhlt2019}
\bibliographystyle{acl_natbib}

\end{document}


\title{\ourmethod: Modeling Higher Order Network Effects in Knowledge Graphs via Network Infused Embeddings}
\author{
Anonymous Submission 6535
}

\title{\ourmethod: Modeling Higher Order Network Effects in Knowledge Graphs via Network infused Embeddings}

\begin{table*}[htb!]
\small
\centering
\begin{subtable}{1.0\textwidth}
	\centering
	\begin{tabular}{|c|c|cc|cc|cc|}
    \textbf{\textsc{MRR}} & \textbf{\textsc{Baseline}} & \multicolumn{2}{c|}{\textbf{\textsc{Baseline+\ourmethod}}} & \multicolumn{2}{c|}{\textbf{\textsc{Increase (Absolute)}}} & \multicolumn{2}{c|}{\textbf{\textsc{Increase (Relative)}}} \\
    \hline
    & & \textbf{\textsc{SHMB}} & \textbf{\textsc{Struct}} & \textbf{\textsc{SHMB}} & \textbf{\textsc{Struct}} & \multicolumn{1}{c}{\textbf{\textsc{SHMB}}} & \multicolumn{1}{c|}{\textbf{\textsc{Struct}} }\\
    \textsc{TransE} & 20.6 & $22.2^{\star}$ & 18.0 & +1.6 & -2.6 & +7.8\% & -12.6\% \\
    \textsc{TransH} & 16.7 & $20.0^{\star}$ & 16.3 & +3.3 & -0.4 & +19.8\% & -2.4\% \\
    \textsc{TransR} & 12.9 & $17.8^{\star}$ & 11.3 & +4.9 & -1.6 & +38.0\% & -12.4\% \\
    \textsc{TransD} & 17.0 & $20.3^{\star}$ & 16.7 & +3.3 & -0.3 & +19.4\% & -1.8\% \\
    \textsc{RESCAL} & 31.3 & $34.7^{\star}$ & 34.5 & +3.4 & +3.2 & +10.9\% & +10.2\% \\
    \textsc{DistMult} & 41.8 & $42.9^{\star}$ & 41.8 & +1.1 & +0 & +2.6\% & +0\% \\
    \textsc{ComplEx} & 42.6 & $43.2^{\star}$ & $43.4^{\star}$ & +0.6 & +0.8 & +1.4\% & +1.9\% \\
    \textsc{Analogy} & 41.8 & $42.8^{\star}$ & 42.2 & +1.0 & +0.4 & +2.4\% & +1.0\%\\
    \hline
    \end{tabular}
    \caption{Mean Reciprocal Rank on WN18-RR }
    \label{tab:mrr_fb}
\end{subtable}
\begin{subtable}{1.0\textwidth}
\centering
	\begin{tabular}{|c|c|cc|cc|cc|}
        \textbf{\textsc{Hits@10}} & \textbf{\textsc{Baseline}} & \multicolumn{2}{c|}{\textbf{\textsc{Baseline+\ourmethod}}} & \multicolumn{2}{c|}{\textbf{\textsc{Increase (Absolute)}}} & \multicolumn{2}{c|}{\textbf{\textsc{Increase (Relative)}}}\\
    \hline
     & & \textbf{\textsc{SHMB}} & \textbf{\textsc{Struct}} & \textbf{\textsc{SHMB}} & \textbf{\textsc{Struct}} & \multicolumn{1}{c}{\textbf{\textsc{SHMB}}} & \multicolumn{1}{c|}{\textbf{\textsc{Struct}} }\\
    \textsc{TransE} & 43.7 & $48.1^{\star}$ & 42.7 & +4.4 & -1.0 & +10.0\% & -2.3\% \\
    \textsc{TransH} & 38.2 & $45.3^{\star}$ & 37.9 & +7.1 & -0.3 & +18.6\% & -0.8\% \\
    \textsc{TransR} & 32.8 & $38.7^{\star}$ & 28.2 & +5.9 & -4.6 & +18.0\% & -14.0\% \\
    \textsc{TransD} & 38.9 & $45.8^{\star}$ & 38.6 & +6.9 & -0.3 & +17.7\% & -0.8\% \\
    \textsc{RESCAL} & 39.9 & $41.8^{\star}$ & 41.2 & +1.9 & +1.3 & +4.8\% & +3.3\% \\
    \textsc{DistMult} & 47.4 & $49.7^{\star}$ & 47.6 & +2.3 & +0.2 & +4.9\% & 0.4\% \\
    \textsc{ComplEx} & 47.6 & $51.2^{\star}$ & $50.0^{\star}$ & +3.6 & +2.4 & +7.5\% & +5.0\% \\
    \textsc{Analogy} & 48.5 & $50.9^{\star}$ & 48.1 & +2.4 & -0.4 & +4.9\% & -0.8\% \\
    \hline
    \end{tabular}
    \caption{Hits@10 Performance on WN18-RR}
    \label{tab:hit@10_fb}
\end{subtable}
\caption{{Performance Improvements of link prediction on several Knowledge Graph embedding baselines when \ourmethod\ is incorporated.}}
\label{tab:main_results}
\end{table*}

\begin{table*}[htb!]
\small
	\centering
	\begin{tabular}{|c|c|cc|cc|}
    \textsc{\textbf{Model}} & \textbf{\textsc{Baseline}} & \textbf{\textsc{+Node2vec}} & \textbf{\textsc{+\ourmethod-shnb}} & \multicolumn{2}{c|}{\makecell{\textbf{\textsc{Improvement}} \\ \textbf{\textsc{\ourmethod\ vs Node2Vec}}}} \\ \hline
    & & & & \textbf{\textsc{Absolute}} & \textbf{\textsc{Relative}}  \\
    \textsc{TransE} & 20.6 & 22.1 & 22.2 & +0.1 & +0.45\% \\
    \textsc{TransH} & 16.7 & 19.7 & $20.0^{\star}$ & +0.3 & +1.52\% \\
    \textsc{TransR} & 12.9 & 17.1 & 17.8 & +0.7 & +4.09\% \\
    \textsc{TransD} & 17.0 & 20.0 & 20.3 & +0.3 & +1.50\% \\
    \textsc{RESCAL} & 31.3 & 34.6 & $34.7^{\star}$ & +0.1 & +0.29\% \\
    \textsc{DistMult} & 41.8 & 42.8 & 42.9 & +0.1 & +0.23\% \\
    \textsc{ComplEx} & 42.6 & 43.1 & 43.2 & +0.1 & +0.23\% \\
    \textsc{Analogy} & 41.8 & 42.8 & 42.8 & +0 & +0\% \\
    \hline
    \end{tabular}
    \caption{Our network embedding method which explicitly incorporates scale generally outperforms Node2Vec.}
     \label{tab_heat_kernel}
\end{table*}
